# Semi-supervised Learning with Density Based Distances

**Avleen S. Bijral**
Toyota Technological Institute
Chicago, IL 60637

**Nathan Ratliff**
Google Inc.
Pittsburg, PA 15206

**Nathan Srebro**
Toyota Technological Institute
Chicago, IL 60637

## Abstract

We present a simple, yet effective, approach to Semi-Supervised Learning. Our approach is based on estimating *density-based distances* (DBD) using a shortest path calculation on a graph. These Graph-DBD estimates can then be used in any distance-based supervised learning method, such as Nearest Neighbor methods and SVMs with RBF kernels. In order to apply the method to very large data sets, we also present a novel algorithm which integrates nearest neighbor computations into the shortest path search and can find exact shortest paths even in extremely large dense graphs. Significant runtime improvement over the commonly used Laplacian regularization method is then shown on a large scale dataset.

## 1 Introduction

The notion of a similarity measure is central to learning algorithms, and many learning methods take as input a distance matrix between data points. The premise behind semi-supervised learning is that the marginal distribution $p(x)$, can be informative about the conditional distribution $p(y|x)$. It is natural, then, to use a distance measure that takes into consideration the marginal distribution. Our approach to semi-supervised learning is to use $p(x)$ to compute more informative distances between unlabeled points and labeled points. These can then be used in any supervised learning method that takes distances between points as inputs.

Denoting the probability density function in $\mathbb{R}^d$ by $f(x)$, we can define a path length measure through $\mathbb{R}^d$ that assigns short lengths to paths through highly density regions and longer lengths to paths through low density regions. We can express such a path length measure as

$$J_f(x_1 \overset{\gamma}{\rightsquigarrow} x_2) = \int_0^1 g\Big(f(\gamma(t))\Big) \, \| \gamma'(t) \|_p \, dt, \quad (1)$$

where $\gamma : [0,1] \to \mathbb{R}^d$ is a continuous path from $\gamma(0) = x_1$ to $\gamma(1) = x_2$ and $g : \mathbb{R}_+ \to \mathbb{R}$ is monotonically decreasing (e.g. $g(u) = 1/u$). Using Equation 1 as a density-based measure of path length, we can now define the *density based distance* (DBD) between any two points $x_1, x_2 \in \mathbb{R}^d$ as the density-based length of a shortest path between the two points

$$D_f(x_1, x_2) = \inf_\gamma J_f(x_1 \overset{\gamma}{\rightsquigarrow} x_2) \quad (2)$$

Figure 1a) depicts the intuition behind these density-based distances. Distances through low-density regions should be much larger than distances through high-density regions. We would say that points **A** and **B** in the figure are more similar, or have a lower density-based distance between them, than points **A** and **C**, even though points **A** and **C** are closer together in terms of their Euclidean distance.

Density-based distances make sense under both the manifold and the cluster assumptions commonly found in the semi-supervised learning literature. When the density is uniform over a manifold, a density-based distances simply measures distances along the manifold (e.g. Figure 1a)), and so using density-based distances essentially corresponds to following the natural geometry of the manifold. When the data appears in distinct high-density clusters separated by low density regions (e.g. as in Figure 1b)), density-based distances within clusters will be low, while density-based distances between points in different clusters will be high.

To exactly compute the density based distance requires not only a search over all possible paths, but also knowledge of the density $p(x)$ and its support, i.e. the data manifold. In a semi-supervised learning scenario we typically do not have direct access to the marginal $p(x)$, but instead have access to large amounts of unlabeled data. We could then use the unlabeled data as a surrogate for the true marginal density and estimate such density-based distances using the unlabeled data points.

Density based distances were first discussed in the context of semi-supervised learning by Sajama and Orlitsky (2005), who suggested estimating the density using kernel density estimation, and then approximating the density-based distances using this estimated density. However, as noted by the authors, kernel density estimation is not very efficient and for high dimensional data one needs a large amount of data to reasonably approximate the true density. This yields poor results on high dimensional data (Sajama & Orlitsky, 2005). In a manifold setting, one can also use a non-linear dimensionality reduction method to estimate the data manifold, and then calculate distances along the manifold (Weinberger & Saul, 2004; Belkin & Niyogi, 2003). Avoiding explicitly recovering the manifold structure leads to a significantly simpler computational method which is less reliant on the specifics of the "manifold assumption" to semi-supervised learning. Not estimating the distances explicitly allows us to handle much higher dimensional data, in which the sample complexity of density estimation is prohibitive.

Alternatively, a simple heuristic was suggested by Vincent and Bengio (2003) in the context of clustering, and is based on constructing a weighted graph over the data set, with weights equal to the squared distances between the endpoints and calculating shortest paths on this graph. In Section 3 we discuss how this relates to the density-based distances defined in (2). In this paper, we investigate this Graph Based Density Based Distance estimate (Graph-DBD), apply it for semi-supervised learning, and extend it to give it more flexibility and allow it to better capture the true density-based distances.

Calculating the Graph-DBD involves a shortest path computation over a large dense graph. For small and medium-sized data sets, this shortest path computation can be done efficiently using Dijkstra's algorithm. However, our problem is fundamentally different from the traditional shortest-path problem in that we impose no restrictions on the set of all valid sequences: our graph is implicitly fully connected— each point in the data set connects to all other points. Traditional search algorithms assume low vertex degree and would therefore scale quadratically with size of the data set in our setting.

To circumvent this problem, we developed a novel variant of Dijkstra's shortest-path algorithm which integrates the nearest-neighbor computation directly into the search and achieves a quasi-linear complexity in the size of the data set whenever nearest neighbor queries can be solved in logarithmic time (Section 4). Our empirical analysis demonstrates substantial practical computational gains using Dijkstra* while retaining provable correctness guarantees. Dijkstra* allows us to use the Graph-DBD for semi-supervised learning on very large data sets.

## 2 Nearest Neighbor and Local Density

A simple intuition serves the analysis in this section. If the distance between nearby neighbors in some neighborhood is small then one can expect the neighborhood to be a high density region. Thus we seek to establish a relation between the nearest neighbor distances and the neighborhood density. This is quantified by the following relationship:

**Theorem 2.1.** *Let $f(x)$ be a smooth density in $\mathbb{R}^d$ and let $x_0 \in \mathbb{R}^d$ be a fixed arbitrary point in the interior of its support. Consider a random sample $X_1, ..., X_n$ drawn i.i.d from $f(X)$. Define the random variable $Z = \min_j ||x_0 - X_j||_p$ measuring the $\ell_p$ distance from $x_0$ to its nearest neighbor in the sample, and denote the median of this random variable as $m_Z$. Then as $n \to \infty$:*

$$f(x_0) = \frac{\ln(2)}{n} \frac{1}{c_{p,d} m_Z^d} + O(\frac{1}{n^2}) \qquad (3)$$

*where $c_{p,d} = 2^d \Gamma(\frac{p+1}{p})^d / \Gamma(\frac{p+d}{p})$ is the volume of the $\ell_p$ unit sphere in $\mathbb{R}^d$.*

*Proof.* For any radius $r$ we have that $\Pr(Z > r) = (1 - p_r)^n$ where $p_r = \Pr(||X - x_0||_p < r) = \int_{||x-x_0||_p < r} f(x)dx$ is the probability mass of the radius $r$ $\ell_p$-ball around $x_0$. Since the density is smooth, for small enough $r$ we have $p_r \to f(x_0) r^d c_{p,d}$. For the median $m_Z$ we thus have $1/2 = \Pr(Z > m_Z) = (1 - f(x_0) m_Z^d c_{p,d})^n$. Solving for the density yields (3). □

Theorem 2.1 suggests using the following nearest-neighbor based estimator for the density:

$$\hat{f}(x) = \frac{\ln(2)}{n} \frac{1}{c_{p,d} Z^d(x)} \qquad (4)$$

where as before $Z(x) = \min_j \| x - X_j \|_p$ is the distance to the nearest neighbor in the sample. Such estimators, based on nearest neighbors, were first proposed by (Loftsgaarden & Quesenberry, 1965). The Theorem establishes that this estimator is asymptotically median-unbiased, i.e.

$$\text{median}(\hat{f}(x)) = f(x) + O(1/n^2). \qquad (5)$$

Unfortunately, although $\hat{f}(x)$ is asymptotically median-unbiased, it is not a consistent estimator and its variances remains constant even as $n \to \infty$. This is demonstrated in Figure 1 c) which shows the median, as well as the 25th and 75th percentiles of $\hat{f}(x_0)$, for the fixed point $x_0$ under a uniform density on the unit square. Although the median does converge very quickly to the true density (namely, 1), the 25th and 75th percentiles remain roughly constant and far from the median, indicating a non-diminishing probability of the estimator being far from the true density.

One way to obtain a consistent estimator is to base the estimator not on the distance to the nearest-neighbor, but rather to the $\log(n)$ nearest-neighbors (Devroye, 1977). Fortunately, this is not necessary for our purposes. As discussed below, we will be satisfied with a constant-variance estimator as our density-based-distance estimator will essentially be an average of many such density estimators.

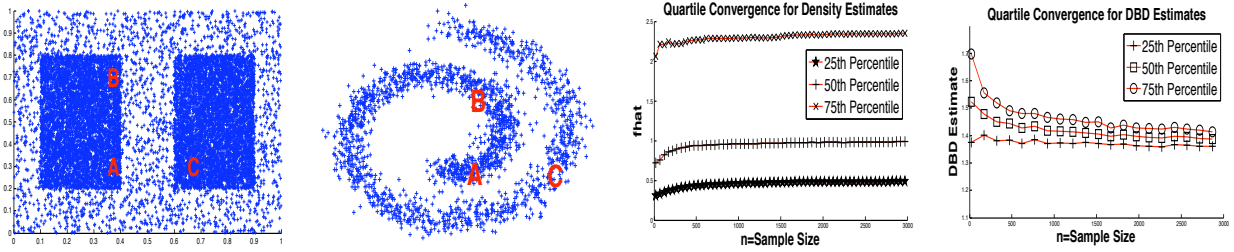

Figure 1: **a)** Cluster Assumption: From a density-based perspective point B is closer to point A than C since they are connected by a path through a region of high density. **b)** Manifold Assumption: point B is closer to point A than C since they are connected by a path through a high density manifold. **c)** 25th, 50th and 75th percentile of the estimator $\hat{f}(x_0)$ at an arbitrary $x_0$. **d)** The Graph Based DBD estimate of the distance between $(0,0)$ and $(1,1)$, for a uniform density over $[0,1]^2$ as a function of the number of points $n$ in the sample. Here $p=2$ and $q=2$ and the DBD was calculated as is described in the text, and then multiplied by $\sqrt{n}$.

## 3 Graph DBDs

The nearest-neighbor based density estimate discussed in the previous Section motivates using the distance between neighboring points as a proxy for the inverse density along the path between them. At least if the points are indeed close together, and are nearest neighbors, or close to being nearest neighbors, the discussion above suggests that the distances between them will indeed be indicative of the inverse density. That is, for points $x_i$ and $x_j$ that are close together, and using $g(f) = 1/f^r$, we can approximate:

$$D_f(x_i, x_j) \approx g(f(\text{around } x_i \text{ and } x_j)) \parallel x_i - x_j \parallel_p$$
$$\propto g(\parallel x_i - x_j \parallel_p^{-d}) \parallel x_i - x_j \parallel_p = \parallel x_i - x_j \parallel_p^{rd+1}$$
$$= \parallel x_i - x_j \parallel_p^q$$

where $q = rd + 1$. Setting $q = 1$ yields the standard, non-density-weighted, $\ell_p$ distance. An important point is that Theorem 2.1 refers to a $d$-dimensional density. Under the manifold assumption, when the data is supported on a low dimensional manifold, there is no density with respect to the ambient dimension, but only a lower dimensional density on the manifold. The dimensionality $d$ in the discussion above should therefore be thought of as the *intrinsic dimensionality*, i.e. the dimensionality of the manifold supporting the data. This dimensionality is typically not known, but luckily in order to apply the proposed method, there is no need to estimate this dimensionality. The method relies on choosing the tuning parameter $q > 1$ (set via cross-validation, or to an arbitrarily chosen moderate value), which can be interpreted as $q = rd + 1$ where $d$ is the intrinsic dimensionality and $r$ is the exponent in $g(f) = 1/f^r$ determining how strongly to weight by the density, but in order to use the method there is no need to tease these two apart.

Returning to approximating the DBD between two further away points, the DBD along a path $x_{\pi(0)} \to x_{\pi(1)} \to x_{\pi(2)} \to \cdots \to x_{\pi(k)}$ hopping between close neighbors could then be approximated as the sum of local DBD estimates along the path:

$$J_f\left(x_{\pi(0)} \overset{\pi}{\leadsto} x_{\pi(k)}\right) = \sum_{i=1}^{k} D_f(x_{\pi(i-1)}, x_{\pi(i)})$$
$$\approx \propto \sum_{i=1}^{k} \parallel x_{\pi(i-1)} - x_{\pi(i)} \parallel_p^q \quad (6)$$

Finally, the density-based-distance between two points could be estimated as the minimum density-based distance over all paths $\pi$ starting at the first point, hopping between other points in the sample, and ending at the other. Note that even though the estimates above were discussed for paths hopping between close-by points, there is no need to restrict the path $\pi$ to only include edges between nearby points: Consider some edge $x_{\pi(i)} \to x_{\pi(i+1)}$ along a path $\pi$. If $x_{\pi(i)}$ and $x_{\pi(i+1)}$ are not close together in the sample, and there are other points in between them, then a path going through these intermediate points will have an estimated DBD based on high density estimates, and would thus have a shorter estimated DBD. Since we defined the DBD in terms of the shortest DBD path, we would prefer a path making short hops over the original path making a long hop.

As discussed in previous Section, using $1/\parallel x_i - x_j \parallel^d$ as a density estimate (up to some global scaling) at each edge is a very rough approximation, with variance that does not shrink with the sample size. However, as the sample size increases, the number of edges on each path increases, with non-neighboring edges being essentially independent. And so, the number of independent contributions to the length of the path grows, and we might expect the variances of its normalized length to decrease. This intuition is confirmed in Figure 1 d) which shows the convergence of estimate of the DBD between two corners of a unit square with uniform density. Further consideration, coping with the dependencies between distances and the minimization over many paths, is necessary in order to solidify this intuition.

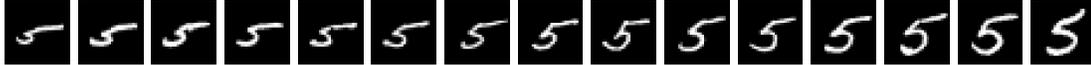

Figure 2: Our algorithm efficiently finds a path from an unlabeled point to the closest labeled point. Although each pair of adjacent points are very similar, the point may change significantly across the entire path. This figure shows an optimal path (of length 15) from query point (left) to labeled point (right) through a data set containing the subset of all MNIST digits 5 and 2 using 10 labeled points from each class (20 in all). We set parameters $p$ and $q$ to 5 and 8, respectively.

In summary, given a sample $x_1, ..., x_n \in \Re^d$, and $\ell_p$ metric, and an exponent $q > 1$, we suggest the following approach to semi-supervised learning:

1. Compute a weight matrix between all pairs of points $x_i$ and $x_j$ (training and test) given by:

$$W_{ij} = \| x_i - x_j \|_p^q . \qquad (7)$$

2. Compute shortest path distances $D_G(i, j)$ between test and the labeled points in a fully connected graph with weights $W_{ij}$.

3. Use a distance-based supervised learning method, such as a Nearest Neighbor method, Parzen Windows method, or Kernel method with a distance-based (e.g. Gaussian) Kernel, with the distances $D_G(i, j)$.

**Comparison With ISOMAP Distances** ISOMAP (J. B. Tenenbaum & Langford, 2000) is a popular method for non-linear dimensionality reduction, which begins with a global distance calculation similar to our Graph Based DBD. ISOMAP distances are also shortest path distances along a graph, but graph edge-lengths are defined as $W_{ij} = \|x_i - x_j\|$ if $x_i, x_j$ are $k$-nearest-neighbors of each other, and infinity (i.e. no edge) otherwise. ISOMAP distances then correspond to geodesic distances along the data manifold, but do not take into consideration varying density: a path through a high density region will have the same length as a path through a low density region. If the data distribution is indeed uniform on a manifold, then with enough points the Graph Based DBD and the ISOMAP distances should be similar. But true distributions are never exactly uniform on a manifold and zero outside the manifold, and so we believe the Graph Based DBD can better capture the relevant distances between points. This is demonstrated on synthetic two-dimensional data in Figure 3, and in experiments later on.

**Computing the Graph-DBD Distances** For small data sets, the matrix $W$ can be computed fully, and a Dijkstra's algorithm can be used in order to perform step 2 above. For larger data sets, where it is impractical to compute the entire matrix $W$, or to invoke Dijkstra on a fully connected graph. One approach is to consider a $k$-nearest-neighbor graph in step 1, thus also reducing the computational cost of Dijkstra's algorithm in step 2. However, this would only be an approximation to the Graph-DBD presented above, and it is not clear a-priori what value of $k$ should be chosen.

It is important to note that unlike methods like ISOMAP that depend delicately on the choice of the number of neighbors, here calculating more neighbors is always beneficial, since the Graph DBD is defined in terms of the *full* graph. In ISOMAP, a too high value of $k$ would cause distances to approach the distances in the ambient space, ignoring the manifold structure, and $k$ is an important parameter controlling in a sense the "resolution" of the manifold structure sought. For Graph-DBD, the exponent $q$ plays a similar role, but if we restrict ourselves to a $k$-nearest-neighbor graph, this is only for computational reasons, and a larger value of $k$ would always be preferable. Of course these would also be more computationally expensive.

Instead of limiting ourselves to such a $k$-nearest-neighbor graph, in the next Section we present a novel algorithm, which we call Dijkstra*, that calculates shortest paths in the fully connected graph, as in the discussion of the Graph-DBD above. However, Dijkstra* only requires access to the few nearest neighbors of each point and is much more computationally efficient than either a full Dijkstra computation on the full graph, or even on a $k$-nearest-neighbor graph with a moderate $k$.

## 4  Dijkstra*: Shortest Paths Through Dense Graphs

The discussions above reduced the problem of semi-supervised learning to a search for shortest-paths in graphs. Our problem, however, differs fundamentally from more traditional search settings in that computing Graph DBDs implies searching through a dense fully-connected data graph. This section introduces a novel variant of Dijkstra's algorithm we call Dijkstra* that integrates the nearest neighbor computation into Dijkstra's inner loop by exploiting a subtle property of the priority queue, thereby allowing it to efficiently search directly through the fully-connected graph. This algorithm obviates the need for a separate $k$ nearest-neighbor graph construction stage to approximate the problem.

### 4.1  An inefficiency in Dijkstra's algorithm

Dijkstra's algorithm is very effective for many shortest path graph search problems. Specifically, the algorithm solves the problem of finding a shortest path from each vertex in the graph to one or more goal vertices. In our case, the set of goals $G$ in our problem is the set of all labeled points in the data set as we would like to find the shortest path between every point and every labeled point. Central to the algorithm is the maintainence of a priority queue containing a collection of candidate paths from the goal. (Typically, implementations represent these paths efficiently as unique paths through a search tree, but there are conceptual advantages to interpreting the contents of the queue as distinct paths.) It is easy to prove that if the queue is initialized to contain the zero-length paths represented by the goals (labeled points) alone, a simple greedy expansion procedure is guaranteed to produce shortest paths through the graph as is described in the paragraphs below.

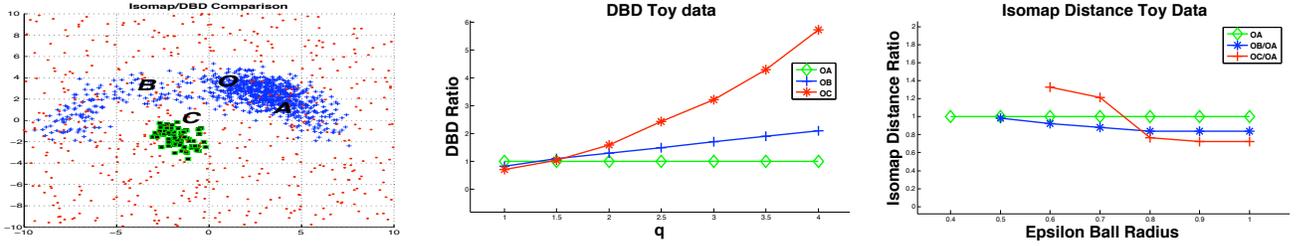

Figure 3: a) Toy example demonstrating the difference between DBD and Isomap distances. For each method, we plot the relative distances between three pairs of points, as we change the parameter $q$ or the radius of the $\epsilon$-ball used to construct the ISOMAP graph.

We will treat the priority queue $Q$ as simply a set of paths where each path in the queue $\xi = (x_0, \ldots, x_l) \in Q$ starts in the goal set ($x_0 \in G$). We can denote the operation of popping the shortest path from the queue as $\xi^* = \arg\min_{\xi \in Q} J(\xi)$, where $J(\xi) = \sum_{i=1}^{l} \|x_{i-1} - x_i\|_p^q$ is simply the path length used by the Graph DBD approximation. (Note that we implicitly assume removal from the queue after each pop operation.) The priority queue data structure (T.H. Cormen & Stein, 2003) provides an efficient representation for $Q$ giving logarithmic complexity to the argmin and set addition operators.

Every iteration of Dijkstra's algorithm pops the current shortest path from the queue $\xi^* = (x_0^*, \ldots, x_l^*) = \arg\min_{\xi \in Q} J(\xi)$. This path is guaranteed to be the shortest path from $x_0^* \in G$ to it's final point $x_l^* \in \mathcal{X}$. The algorithm then adds a collection of new paths to the queue of the form $(\xi^*, x_{l+1})$, where $x_{l+1}$ ranges over all neighbors $\mathcal{N}(x_l^*)$ of the final point $x_l^*$. Adding this collection to the queue maintains the optimality guarantee of subsequent iterations (T.H. Cormen & Stein, 2003).

Unfortunately, for our problem, since we search through fully connected graphs, the expansion step of Dijkstra's algorithm can be prohibitively expensive: at each iteration, the algorithm would add on the order of $|\mathcal{X}|$ new paths to the priority queue. A simple intuition, however, underscores an inefficiency in this approach, which we exploit in deriving Dijkstra*: although Dijkstra adds all neighboring paths to the queue during a given iteration, it is always the single shortest of these paths that is the first to be popped if

**Algorithm 1** Dijkstra* search
1: initialize $Q = \{(x_0)\}_{x_0 \in \mathcal{G}}$.
2: **while** $|Q| > 0$ **do**
3:     pop from queue,
        $\xi^* = (x_0, \ldots, x_T) \leftarrow \arg\min_{\xi \in Q} J(\xi)$
        and remove $x_T$ from $\mathcal{X}$
4:     push_next($Q, \xi^*$)
5:     if $T > 0$ then push_next($Q, \xi_{T-1}^*$)
6: **end while**

**Algorithm 2** push_next($Q, \xi = (x_0, \ldots, x_T)$)
1: compute nearest-neighbor
    $x' = \arg\min_{x \in \mathcal{X}} \|x - x_T\|_p$
2: add extended path $\xi' = (x_0, \ldots, x_T, x')$ to $Q$

any are popped at all. We can therefore just add the single best neighboring path, as long as we systematically add the next best path once that path is popped (if it ever is).

### 4.2 The Dijkstra* algoritm

Following the intuition of the preceeding section, we suggest a novel variant of Dijkstra's algorithm called Dijkstra* designed to avoid expanding all neighbors of a given vertex simultaneously, as the degree of each vertex in our setting is very large.[1] Algorithm 1 lists pseudocode for Dijkstra*, and Theorem 4.1 below proves its correctness (proof not given because of lack of space)

**Theorem 4.1.** *If $\xi = (x_0, \ldots, x_T)$ is popped from $Q$ during Step 3 of Algorithm 1, then $\xi$ is an optimal path from $x_0$ to $x_T \in \mathcal{S}$. Moreover, each node $x \in \mathcal{X}$ will eventually be the end point of some popped path $\xi$.*

### 4.3 Achieving $O(n \log n)$ using space partitioning trees

In the worse case, nearest neighbor queries are fundamentally hard and take $O(n)$ time to perform. However, when the geometric structure of the data is nice (e.g. low intrinsic dimensionality) space partitioning trees such as the Cover Tree have proven effective in reducing that complexity to $O(\log n)$ both theoretically and practically (Beygelzimer & Langford, 2006). Dijkstra* can easily use these data structures to improve its inner loop complexity. Denoting the total number of vertices and edges in a graph as $V$ and $E$, respectively, the generic running time of Dijkstra's traditional algorithm is $O((V + E)\log V)$ when the priority queue is implemented using a binary heap (T.H. Cormen & Stein, 2003). For our problem, however, $E = O(V^2)$ and $V = n$ (the number of data points) so the complexity is expressed more precisely as $O(n^2 \log n) = \tilde{O}(n^2)$. When $n$ is large, this quadratic complexity is computationally infeasible. However, when space partitioning trees achieve $O(\log n)$ performance on the nearest neighbor query and point removal operations, the $O(n)$ inner loop complexity of Dijkstra* vanishes and the overall complexity reduces to $O(n \log n)$.

---
[1] In the tradition of A*, we interpret the * in Dijkstra*'s name as the UNIX wildcard pattern denoting "all versions".

| $k$ | 15 | 20 | 30 | 50 | 100 | speedup |
|---|---|---|---|---|---|---|
| *Dijkstra\** | 2.429 | 2.929 | 3.745 | 5.206 | 10.567 | - |
| *Dijkstra* | 4.144 | 5.976 | 10.023 | 15.410 | 52.586 | 5x |
| *Laplacian Reg.* | 18.484 | 24.25 | 35.581 | 57.67 | 117.79 | 11x |
| *unreachable* | 2335 | 1757 | 988 | 621 | 527 | - |

Table 1: Timing results in seconds for semi-supervised classification of the CoverType data set using $k$-nearest-neighbor graphs with $k$ ranging from 15 through 100 using 100 labeled points. Each values is the average over 10 different independent samplings of the 100 labeled points. The final row depicts the average number of points that were residing in components disconnected from the labeled set as $k$ varies. The final column shows the multiplicative speedup of Dijkstra* over both alternative algorithms for $k = 100$.

| Algorithm | $k=15$ | $k=20$ | $k=30$ | $k=50$ | $k=100$ |
|---|---|---|---|---|---|
| 1-NN Isomap | 0.44079 | 0.445 | 0.45351 | 0.46539 | 0.47442 |
| 1-NN DBD(p=2,q=4) | 0.4345 | 0.43446 | 0.43446 | 0.43453 | 0.43439 |
| Laplacian Reg. | 0.43194 | 0.43246 | 0.43307 | 0.43433 | 0.43458 |
| 1-NN Euclidean | 0.5175 | 0.5175 | 0.5175 | 0.5175 | 0.5175 |

Table 3: Quantitative average classification results on the CoverType data set. The results (error rates) are averaged over 10 randomly chosen subsets of 100 labeled and the rest unlabeled points. The algorithms are run on these subsets for varying $k$ for the underlying nearest neighbor graphs.

Note that additionally, although we always need to compute nearest neighbors, by integrating this computation directly into Dijkstra*, we avoid arbitrarily choosing a single global vertex degree $k$ across the entire data set. Dijkstra* more efficiently uses each nearest neighbor query by avoiding previously closed nodes (see Figure 4 and its caption for details).

## 5 Experiments

We divide the experiments into two sets, one studying the medium sized benchmarks detailed in (O. Chapelle & Zien, 2006) and the other addressing the larger scale CoverType and MNIST datasets. We show comparable and sometimes better accuracy on large data sets while achieving far superior runtime performance.

### 5.1 Medium-Size Benchmarks

In this set of experiments we show results on the semi-supervised learning benchmarks described in (O. Chapelle & Zien, 2006). We used the datasets Digit11, USPS, COIL, BCI and TEXT. The first four are described as "manifold-like" and are assumed to lie on a low dimensional manifold. The last dataset is described as "cluster-like" and assumed to contain label-homogeneous clusters. The datasets can be obtained at (Data, 2006).

The methods against which we compare have also been described in the above reference and the results for these algorithms are reproduced from that publication. These algorithms include manifold and cluster based algorithms. Please see (O. Chapelle & Zien, 2006) for more details on the data and the algorithms.

Table 2 shows the results on the benchmark datasets. For the 10 labeled point case the results reported are the best obtained on the test set. In the 100 labeled point case the parameters $p$ and $q$ were obtained using 10-fold cross validation over each of the 12 labeled subset of points and a subset of the unlabeled points. Thus for each subset we have a parameter pair $(p, q)$. A simple nearest neighbor classifier using Graph DBDs is very competitive with the state-of-the-art while being significantly more computationally efficient than the competing algorithms.

We also show results for the fixed values of $p = 2$ and $q = 8$ in the last row. These values were chosen arbitrarily and represent a reasonable choice for most datasets. Most of the best performing methods on the manifold-like data involves very heavy computations on the graph Laplacian including computing eigenvectors and solving linear equations (O. Chapelle & Zien, 2006). In comparison our Graph DBD computation requires only a distance matrix and a shortest path calculation. The choice of the parameters $p$ and $q$ also seems to be a relatively easy problem since for most of the datasets we the only choices that seem to work well lie between 2 and 8.

Throughout these experiments we used a very simple 1-nearest-neighbor classifier using our Graph DBDs. More sophisticated distance-based supervised learning methods may lead to better results.

### 5.2 Large Scale Experiments

In the final set of experiments we used the MNIST ($n = 60000, d = 784$) and the CoverType ($n = 581012, d = 54$) datasets. For both datasets we average results over 10 random splits of labeled and unlabeled points each.

For datasets of these sizes we would ideally integrate an efficient nearest-neighbor data structure directly into the Dijkstra* algorithm. Unfortunately, such algorithms are known to achieve only approximately lin-ear query performance on the MNIST data set, and our cover tree implementation was substantially slower than the code provided by the authors of (Beygelzimer & Langford, 2006), which implements only batch nearest-neighbor operations. We therefore compute the Graph DBD between points using $K$-NN graphs in both cases. For the MNIST data set, we computed nearest-neighbors using a simple brute force computation, and for the CoverType data set, we used the above mentioned code provided by the authors.

| Supervised | Digit1 | USPS | COIL | BCI | Text | | Supervised | Digit1 | USPS | COIL | BCI | Text |
|---|---|---|---|---|---|---|---|---|---|---|---|---|
| 1-NN | 23.47 | 19.82 | 65.91 | 48.74 | 39.44 | | 1-NN | 6.12 | 7.64 | 23.27 | 44.83 | 30.77 |
| SVM | 30.6 | 20.03 | 68.36 | 49.85 | 45.37 | | SVM | 5.53 | 9.75 | 22.93 | 34.31 | 26.45 |
| *Semi-supervised* | | | | | | | *Semi-supervised* | | | | | |
| MVU+1-NN | 11.92 | 14.88 | 65.72 | 50.24 | 39.4 | | MVU+1-NN | 3.99 | 6.09 | 32.27 | 47.42 | 30.74 |
| LEM+1-NN | 12.04 | 19.14 | 67.96 | 49.94 | 40.84 | | LEM+1-NN | *2.52* | 6.09 | 36.49 | 48.64 | 30.92 |
| QC+CMN | 9.8 | *13.61* | 59.63 | 50.36 | 40.79 | | QC+CMN | 3.15 | 6.36 | 10.03 | 46.22 | 25.71 |
| Discrete Reg. | 12.64 | 16.07 | 63.38 | 49.51 | 40.37 | | Discrete Reg. | 2.77 | *4.68* | *9.61* | 47.67 | 24 |
| SGT | 8.92 | 25.36 | - | 49.59 | 29.02 | | SGT | 2.61 | 6.8 | - | 45.03 | 23.09 |
| LDS | 15.63 | 17.57 | 61.9 | 49.27 | 27.15 | | LDS | 3.46 | 4.96 | 13.72 | 43.97 | 23.15 |
| Laplacian RLS | *5.44* | 18.99 | (*54.54*) | 48.97 | 33.68 | | Laplacian RLS | 2.92 | *4.68* | (11.92) | *31.36* | 23.57 |
| CHM (normed) | 14.86 | 20.53 | - | *46.9* | - | | CHM (normed) | 3.79 | 7.65 | - | 36.03 | - |
| TSVM | 17.77 | 25.2 | 67.5 | 49.15 | 31.21 | | TSVM | 6.15 | 9.77 | 25.8 | 33.25 | 24.52 |
| Cluster-Kernel | 18.73 | 19.41 | 67.32 | 48.31 | 42.72 | | Cluster-Kernel | 3.79 | 9.68 | 21.99 | 35.17 | 24.38 |
| Data-Dep. Reg. | 12.49 | 17.96 | 63.65 | 50.21 | - | | Data-Dep. Reg. | 2.44 | 5.1 | 11.46 | 47.47 | - |
| 1-NN(DBD) | (11.06) | (14.24) | (59.40) | (48.44) | (37.25) | | 1-NN(DBD) | 4.80 | 6.55 | 9.65 | 44.86 | 28.7 |
| 1-NN(DBD)(p=2,q=8) | 14.838 | 15.57 | 60.37 | 49.06 | 37.20 | | 1-NN(DBD)(p=2,q=8) | 4.75 | 6.64 | 9.98 | 45.61 | 28.7 |

Table 2: Test Errors (%) with 10 and 100 labeled points respectively. The best performing algorithm is italicized and in bold. Model selection for results in parenthesis was performed with respect to the test set.

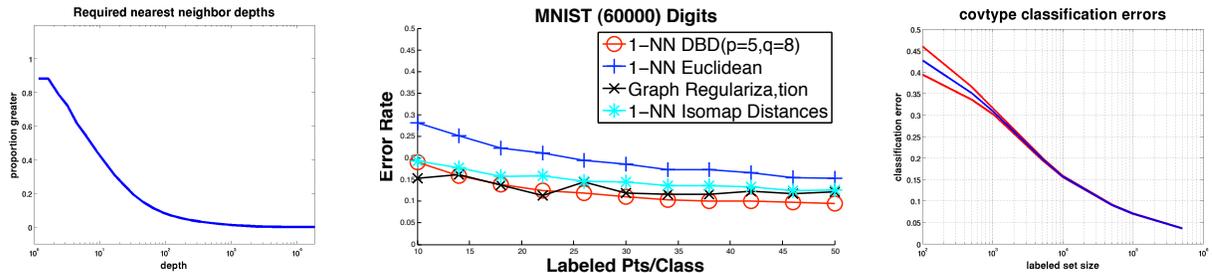

Figure 4: **Left:** Dijkstra* makes on average only 7 to 8 nearest neighbor queries per node. However, since these queries explicitly avoid all previously closed nodes, the effective "depth" of the query is much larger. This plot, computed using the MNIST data set, shows the proportion of edges along optimal paths whose true nearest neighbor depths exceeds a particular value. **Middle:** Error rates for the entire MNIST training set with 60000 digits with the results averaged over 10 splits of labeled/unlabeled. 1-NN dbd outperforms both 1-NN Isomap and Graph regularization. **right:** Results of DBD nearest neighbor classification on the CoverType data set.

**MNIST.** For the MNIST data set, we arbitrarily set $K$ to 200. and compared the performance of our Graph DBD approach (1-NN DBD) to 1-NN classification under both Isomap and Euclidean distances (1-NN Isomap and 1-NN Euclidean, resp.), and to the Graph Regularization approach of (Zhou & Bousquet, 2003). In this experiment, we varied the number of labeled points from 10 points per class (100 total) through 50 points per class (500 total). The parameters $p, q$ for our DBD and the parameter $\sigma$ for the Gaussian kernel in the Graph regularization method were chosen by cross-validation on a labeled set of size 100. The validation was performed over 5 random splits of 40 labeled and 60 unlabeled points. Figure 4 depicts these results. Our 1-NN DBD algorithm generally outperforms all other methods, except for graph regularization on the smallest labeled set sizes.

**CoverType** For our final experiment, we ran our algorithm on the CoverType data set consisting of over half a million points. The experiments detailed here demonstrate that our algorithm is effective and efficient even on very large data sets.

We compare our method against 1-NN classifiers using both Isomap and Euclidean distances and to the Graph regularization approach. We fixed $p = 2$ in the paper for our DBD. The parameter for the Gaussian kernel used in the Graph regularization algorithm and the value of $q$ for our method were tuned manually on the labeled dataset.

We show the results for different $K$ values used to construct the underlying nearest neighbor graph. We can see in Table 3 that we perform almost as well as Graph regularization and generally outperform both competing 1-NN classification variants (under Isomap and Euclidean distances). However, as depicted by the runtime statistics of Table 1, the Graph Regularization method (where we use Matlab's large scale sparse linear solver) takes increasingly long as $K$ increases. The method will clearly become infeasible if we were to assume a dense graph since it will have to solve a huge (roughly half million times half million) dense linear equation. Our method does not suffer from these limitations as our timing results show, even with a fixed $K$-NN graph we see an improvement over the traditional Dijkstra algorithm simply because it avoids adding extraneous nodes to the queue that will never be popped.

Note that 1-NN Euclidean results obviously do not depend on the underlying graph.

Moreover, the rightmost subplot of Figure 4 shows that our approach continuously transitions from a efficient competitive semi-supervised learning technique to a high-performing supervised learning algorithm as the number of labeled points $l$ increases. At $l = 100$ and $l = 500$ this problem is squarely within the class of semi-supervised learning and we achieve errors of 42.7 and 35.0, respectively. However, when $l$ is increased to $100,000$ and $500,000$, we achieve errors of 7.0 and 3.6, which are competitive with the state-of-the-art in supervised classification for this data set. For each setting of $l$, we sample over 10 randomly chosen labeled sets. The plot gives the average accuracy in blue and the standard deviation error bard in red. For this experiment we chose $p = 2$ and $q = 8$ arbitrarily without attempting to optimize the parameters.

## 6 Conclusions

We present a simple and computationally light approach to semi-supervised learning: estimate density-based distances using shortest path calculations on a graph containing all labeled and unlabeled points, then use these distances in a supervised distance-based learning method. These shortest path distance approximations can be computed efficiently even through a dense fully connected graph. We presented experimental results on benchmark datasets demonstrating that this simple approach, even when combined with a simple 1-nearest-neighbor classifier, is often competitive with much more computationally intensive methods. In particular, the prediction results are often better than those obtained using other, more complex, methods for learning distances using unlabeled data, e.g. by explicitly learning the data manifold.

A point we would like to emphasize, and we hope was made clear by the experimental results, is that semi-supervised learning does not require explicit estimation of neither the data manifold nor the density, and estimating these objects is often more complicated then just calculating density-based distances for use in semi-supervised learning. Moreover, our algorithm is fairly robust to the choice of parameters. In the experiments on the benchmark datasets we tried using $q = 8$ for all data-sets and all sample sizes, generally obtaining good results.